\documentclass[twoside,11pt]{article}

\usepackage{jmlr2e}

\usepackage{amsfonts, amsmath, amssymb}
\usepackage{dblfnote}
\DFNalwaysdouble 
\usepackage{enumitem}
\usepackage{fontawesome} 
\usepackage{graphicx}
\usepackage{subfigure}

\usepackage{hyperref}
\hypersetup{
    breaklinks=true,
    colorlinks=true,
    linkcolor=mydarkblue,
    citecolor=mydarkblue,
    filecolor=mydarkblue,
    urlcolor=mydarkblue
}
\usepackage[utf8x]{inputenc}
\usepackage{listings}
\usepackage{pgffor}
\usepackage{natbib}

\usepackage[usenames, dvipsnames, svgnames]{xcolor}
\definecolor{mydarkblue}{rgb}{0,0.08,0.45}

\usepackage{xargs} 

\newcommand{\DPPy}{\textsf{DPPy}}
\newcommand\DPP{\operatorname{DPP}}

\newcommand*{\Eqref}[1]{Equation\,\ref{#1}}

\newcommand{\dist}{\operatorname{distance}}

\newcommand{\CommaBin}{\mathbin{\raisebox{0.5ex}{,}}}

\newcommandx{\mycitet}[3][1=, 2=]{\citeauthor{#3}\,\citeyearpar[#1][#2]{#3}}

\newcommand\eg{\text{e.g., }}
\newcommand\iid{\text{i.i.d.\,}}

\newcommand{\footGitHubDPPy}{
\href{https://github.com/guilgautier/DPPy}
     {\textsf{github.com/guilgautier/DPPy}}
}

\newcommand{\footReadTheDocs}{
\href{https://dppy.readthedocs.io}
     {\textsf{dppy.readthedocs.io}}
}

\newcommand{\footTravis}{
\href{https://travis-ci.com/guilgautier/DPPy}
     {\textsf{travis-ci.com/guilgautier/DPPy}}
}

\newcommand{\footCoveralls}{
\href{https://coveralls.io/github/guilgautier/DPPy}
     {\textsf{coveralls.io/github/guilgautier/DPPy}}
}

\makeatletter
\@addtoreset{footnote}{page}
\makeatother

\renewcommand{\thefootnote}{
    \ifcase\value{footnote}
        \or{1}
        \or{2}
        \or{3}
        \or{\color{black}\faGithub}
        \or{\color{black}\faBook}
        \or{\color{black}\faGears}
        \or{\color{black}\faIndent}
        \or{\color{black}\faNewspaperO}
    \fi}

\makeatletter
\newcommand\footnoteref[1]{\protected@xdef\@thefnmark{\ref{#1}}\@footnotemark}
\makeatother

\renewcommand{\tilde}{\widetilde}

\newcommand{\suml}{ \sum\limits }

\foreach \x in {A,...,Z}
	{%
	\expandafter\xdef\csname bf\x \endcsname{\noexpand\ensuremath{\noexpand\mathbf{\x}}}
	\expandafter\xdef\csname cal\x \endcsname{\noexpand\ensuremath{\noexpand\mathcal{\x}}}
	\expandafter\xdef\csname bb\x \endcsname{\noexpand\ensuremath{\noexpand\mathbb{\x}}}
}

\newcommand*\diff{\mathop{}\!\mathrm{d}}

\newcommand{\lrb}[1]{\left[ #1 \right]}
\newcommand{\lrp}[1]{\left( #1 \right)}
\newcommand{\lrcb}[1]{\left\{ #1 \right\}}
\newcommand{\lrabs}[1]{\left| #1 \right|}

\newcommand\Prob{\operatorname{\bbP}}
\newcommand\Exp{\operatorname{\bbE}}

\newcommand{\Proba}[1]{\Prob\!\lrb{#1}}
\newcommand{\Expe}[1]{\Exp\!\lrb{#1}}

\newcommand\Ber{\operatorname{\calB er }}

\renewcommand{\top}{\mathsf{\scriptscriptstyle T}}

\newcommand\Span{\operatorname{span}}

\newcommand\rank{\operatorname{rank}}

\definecolor{white}{rgb}{1,1,1}
\definecolor{light_gray}{rgb}{0.97,0.97,0.97}
\definecolor{mykey}{rgb}{0.117,0.403,0.713}
\definecolor{myviolet}{hsb}{0.79167,1,1}

\makeatletter
\newcommand{\ProcessDigit}[1]
{%
  \ifnum\lst@mode=\lst@Pmode\relax%
   {\color{OliveGreen} #1}%
  \else
    #1%
  \fi
}
\makeatother

\lstdefinelanguage{mypython}{
    aboveskip=1mm,
    belowskip=1mm,
    columns=flexible,
    mathescape,
    showstringspaces=false,
    basicstyle=\ttfamily\footnotesize,
    keywordstyle=\bfseries\color{OliveGreen},
    morekeywords={access,and,assert,break,class,continue,def,del,elif,else,%
    except,exec,False,finally,for,from,global,if,import,in,is,lambda,not,raise,return,True},
    morekeywords={[2] abs,all,any,basestring,bin,bool,bytearray,callable,chr,
    classmethod,cmp,compile,complex,delattr,dict,dir,divmod,enumerate,eval,
    execfile,file,filter,float,format,frozenset,getattr,globals,hasattr,hash,
    help,hex,id,input,int,isinstance,issubclass,iter,len,list,locals,long,map,
    max,memoryview,min,next,object,oct,open,ord,pow,property,print,range,raw_input,
    reduce,reload,repr,reversed,round,set,setattr,slice,sorted,staticmethod,str,sum,super,tuple,type,unichr,unicode,vars,xrange,zip,apply,buffer,coerce,
    intern},
    keywordstyle=[2]\color{OliveGreen},
    stringstyle=\ttfamily\color{FireBrick},
    morestring=[b]{'},
    morestring=[b]{"},
    morestring=[s]{r'}{'},
    morestring=[s]{r"}{"},
    morestring=[s]{r'''}{'''},
    morestring=[s]{r"""}{"""},
    morestring=[s]{u'}{'},
    morestring=[s]{u"}{"},
    morestring=[s]{u'''}{'''},
    morestring=[s]{u"""}{"""},
    commentstyle=\slshape\textcolor{CadetBlue},
    morecomment=[l]\#,%
    morecomment=[s]{'''}{'''},
    morecomment=[s]{"""}{"""},
    literate=*%
        {0}{{{\ProcessDigit{0}}}}1
        {1}{{{\ProcessDigit{1}}}}1
        {2}{{{\ProcessDigit{2}}}}1
        {3}{{{\ProcessDigit{3}}}}1
        {4}{{{\ProcessDigit{4}}}}1
        {5}{{{\ProcessDigit{5}}}}1
        {6}{{{\ProcessDigit{6}}}}1
        {7}{{{\ProcessDigit{7}}}}1
        {8}{{{\ProcessDigit{8}}}}1
        {9}{{{\ProcessDigit{9}}}}1
        {:}{{\bfseries:}}2%
        {::}{{\bfseries:$\!$:}}2%
        {=}{{\bfseries\color{myviolet}{=}}}2%
        {-}{{\bfseries\color{myviolet}-}}{2}%
        {+}{{\bfseries\color{myviolet}+}}{2}%
        {*}{{\bfseries\color{myviolet}*}}2%
        {**}{{\bfseries\color{myviolet}{**}}}3%
        {/}{{\bfseries\color{myviolet}/}}{2}%
        {//}{{\bfseries\color{myviolet}{//}}}{2}%
        {!}{{\bfseries\color{myviolet}!}}{2}%
        {<}{{\bfseries\color{myviolet}<}}{2}%
        {<=}{{\bfseries\color{myviolet}{<=}}}3%
        {>}{{\bfseries\color{myviolet}>}}{2}%
        {>=}{{\bfseries\color{myviolet}{>=}}}3%
        {==}{{\bfseries\color{myviolet}{==}}}3%
        {!=}{{\bfseries\color{myviolet}{!=}}}3%
        {+=}{{\bfseries\color{myviolet}{+=}}}3%
        {-=}{{\bfseries\color{myviolet}{-=}}}3%
        {*=}{{\bfseries\color{myviolet}{*=}}}3%
        {/=}{{\bfseries\color{myviolet}{/=}}}3,
    morekeywords={[3] jacobian_points_weights_to_moments_circle},
    keywordstyle=[3]\color{blue}
}

\jmlrheading{xx}{2019}{xx-xx}{8/12}{xx/xx}{gabava18}{Gautier,
                                                     Guillermo Polito,
                                                     Rémi Bardenet
                                                     and Michal Valko}

\ShortHeadings{\DPPy}{Gautier,
                      Polito,
                      Bardenet
                      and Valko}
\firstpageno{1}

\begin{document}

\title{\DPPy: Sampling DPPs with Python}

\author{\name Guillaume Gautier$^{\dagger*}$
            \email g.gautier@inria.fr\\
        \name Guillermo Polito$^{\dagger*}$
            \email guillermo.polito@univ-lille.fr\\
        \name R\'emi Bardenet$^\dagger$
            \email remi.bardenet@gmail.com\\
        \name Michal Valko$^{*\ddag}$
            \email valkom@deepmind.com\\
        \addr $^\dagger$Univ.\,Lille, CNRS, Centrale Lille, UMR 9189\,--\,CRIStAL,  59651 Villeneuve d'Ascq, France\\
        \addr $^*$Inria Lille-Nord Europe, 40 avenue Halley 59650 Villeneuve    d'Ascq, France\\
        \addr $^\ddag$DeepMind Paris, 14 Rue de Londres, 75009  Paris, France
}

\editor{}

\maketitle

\vspace{-3em}
\setcounter{footnote}{3}
\begin{abstract}
    Determinantal point processes (DPPs) are specific probability distributions over clouds of points that are used as models and computational tools across physics, probability, statistics, and more recently machine learning.
    Sampling from DPPs is a challenge and therefore we present \DPPy, a Python toolbox that gathers known exact and approximate sampling algorithms for both finite and continuous DPPs.
    The project is hosted on GitHub\!\footnote{\label{fn:github}\footGitHubDPPy}and equipped with an extensive documentation.\!\!\footnote{\label{fn:docs}\footReadTheDocs}
\end{abstract}


\vspace{-1em}

\section{Introduction} 
\label{sec:introduction}

    Determinantal point processes (DPPs) are distributions over configurations of points that encode diversity through a kernel function $K$.
    They were introduced by \citet{Mac75} as models for beams of fermions, and they have since found applications in fields as diverse as probability \citep{Sos00, Kon05, HKPV06}, statistical physics \citep{PaBe11}, Monte Carlo methods \citep{BaHa16}, spatial statistics \citep{LaMoRu15}, and machine learning \citep[ML,][]{KuTa12}.

    In ML, DPPs mainly serve to model diverse sets of items, as in recommendation \citep{KaDeKo16, GaPaKo16} or text summarization \citep{DuBa18}.
    Consequently, MLers  use mostly finite DPPs, which are distributions over subsets of a finite \emph{ground set} of cardinality $M$, parametrized by an $M\times M$ kernel matrix $\bfK$.
    Routine inference tasks such as normalization, marginalization, or sampling have complexity $\calO(M^3)$ \citep{Gil14}.
    Like other kernel methods, when $M$ is large, $\calO(M^3)$ is a bottleneck.

    In terms of software, the R library \textsf{spatstat} \citep{BaTu05}, a general-purpose toolbox on spatial point processes, includes sampling and learning of continuous DPPs with stationary kernels, as described by \mycitet{LaMoRu15}.
    Complementarily, we propose \DPPy, a turnkey Python implementation of known general algorithms to sample \emph{finite} DPPs.
    We also include algorithms for non-stationary continuous DPPs, e.g., related to random covariance matrices or Monte Carlo methods that are also of interest for MLers.

    The \DPPy\ project, hosted on GitHub,\!\footnoteref{fn:github}is already being used by the cross-disciplinary DPP community (\citealp{BuRaWi19,Kam18,Pou19,DeCaVa19,GaBaVa19}).
    We use Travis\!\footnote{\footTravis}for continuous integration and Coveralls\!\footnote{\footCoveralls}for test coverage.
    Through ReadTheDocs\footnoteref{fn:docs}we provide an extensive documentation, which covers the essential mathematical background and showcases the key properties of DPPs through \DPPy\ objects and associated methods.
    \DPPy\ thus also serves as a tutorial.




    \section{Definitions} 
    \label{sec:definitions}

        A point process is a random subset of points $\calX=\lrcb{X_1, \dots, X_N} \subset \bbX$, where the number of points $N$ is itself random.
        We further add to the definition that $N$ should be almost surely finite and that all points in a sample are distinct.
        Given a reference measure $\mu$ on~$\bbX$, a point process is usually characterized by its $k$-correlation function $\rho_k$ for all $k$, where
        \begin{equation*}
        \label{eq:correlation_function_intuition}
            \Proba{
                \begin{tabular}{c}
                    $\exists$ one point of the process in\\
                    each ball $B(x_i, \diff x_i), \forall i=1,\dots, k $
                \end{tabular}
            }
            = \rho_k\lrp{x_1,\dots,x_k}
                \prod_{i=1}^k \mu(\diff x_i),
        \end{equation*}
        see \citet[][Section\,4]{MoWa04}.
        The functions $\rho_k$ describe the interaction among points in $\calX$ by quantifying co-occurrence of points at a set of locations.


        A point process $\calX$ on $(\bbX,\mu)$ parametrized by a kernel $K:\bbX\times \bbX\rightarrow \bbC$ is said to be \emph{determinantal}, denoted as $\calX\sim\DPP(K)$, if its $k$-correlation functions satisfy
        \begin{equation*}
            \label{eq:k-correlation_function_DPP}
            \rho_k(x_1,\dots,x_k)
              = \det \lrb{K(x_i, x_j)}_{i,j=1}^k,
            \quad \forall k\geq 1.
        \end{equation*}
        In ML, most DPPs are in the finite setting where $\bbX = \lrcb{1,\dots,M}$ and $\mu=\sum_{i=1}^M \delta_i$.
        In this context, the kernel function becomes an $M\times M$ matrix $\bfK$, and the correlation functions refer to inclusion probabilities.~DPPs are thus often defined as $\calX\sim \DPP(\bfK)$ if
        \begin{equation}
        \label{eq:inclusion_proba_finite}
            \Proba{S \subset \calX} = \det \bfK_S,
                \quad\forall S\subset \bbX,
        \end{equation}
        where ${\bfK}_S$ denotes the submatrix of $\bfK$ formed by the rows and columns indexed by $S$.
        The kernel matrix $\bfK$ is commonly assumed to be real-symmetric, in which case the existence and uniqueness of the DPP in \Eqref{eq:inclusion_proba_finite} is equivalent to the condition that the eigenvalues of $\bfK$ lie in $[0,1]$.
        The result also holds for general Hermitian kernel functions $K$ with additional assumptions \cite[Theorem\,3]{Sos00}.
        We note that there are also DPPs with nonsymmetric kernels \citep{BoDiFu10,GBDK19}.

        Oftentimes, ML practitioners favor a more flexible definition of a DPP in terms of a \emph{likelihood kernel} $\bfL$, which only requires $\bfL\succeq 0$ so that
        \begin{equation*}
            \Proba{\calX=S} = \frac{\det \bfL_S}{\det \lrb{I + \bfL}}\CommaBin
        \end{equation*}
        rather than a \emph{correlation kernel} $0\preceq \bfK \preceq I$.
        Yet, the $\bfL$ parametrization makes \Eqref{eq:inclusion_proba_finite} less interpretable and does not cover important cases such as fixed size DPPs which are achievable using projection $\bfK$ kernels.
        \citet[][Section\,5]{KuTa12} countered that with $k$-DPPs, which can be understood as DPPs parametrized by a likelihood kernel, conditioned to have exactly $k$ elements.
        However, in general, $k$-DPPs are not DPPs.

        The main interest in DPPs in ML is that they model diversity while being tractable.
        Compared to independent sampling with the same marginals, \Eqref{eq:inclusion_proba_finite} entails
        \begin{equation*}
        \label{eq:2point_correlation_function_inclusion_proba_finite_case}
          \Proba{\{i,j\} \subset \calX}
            = {\bfK}_{ii}{\bfK}_{jj}-{\bfK}_{ij}{\bfK}_{ji}
            = \Proba{\{i\} \subset \calX}
              \times \Proba{\{j\} \subset \calX}
                - |{\bfK}_{ij}|^2,
        \end{equation*}
        so that, the larger $\lrabs{\bfK_{ij}}$ less likely items $i$ and $j$ co-occur.
        If $\bfK_{ij}$ models the similarity between items $i$ and $j$, DPPs are thus random \emph{diverse} sets of elements.

        Most point processes that encode diversity are not tractable, in the sense that efficient algorithms to sample, marginalize, or compute normalization constants are not available.
        However, DPPs are amenable to these tasks with polynomial complexity \citep{Gil14}.
        Next, we present the challenging task of sampling, which is the core of \DPPy.


    \section{Sampling determinantal point processes} 
    \label{sec:sampling}

        We assume henceforth that $K$ is real-symmetric and satisfies suitable conditions \citet[][Theorem\,3]{Sos00} so that its spectral decomposition is available
        \begin{equation*}
        \setlength{\belowdisplayskip}{0pt}
        \setlength{\belowdisplayshortskip}{0pt}
        \setlength{\abovedisplayskip}{0pt}
        \setlength{\abovedisplayshortskip}{0pt}
        \label{eq:eigdec_kernel}
        K(x,y)
        \triangleq
          \suml_{i=1}^{\infty}
            \lambda_i \phi_i(x) \phi_i(y),
          \quad \text{with }
            \int_{\bbX} \phi_i(x) \phi_j(x) \mu(\diff x) = \delta_{ij}.
        \end{equation*}
        Note that, in the finite case, the spectral theorem is enough to eigendecompose $\bfK$.
        \mycitet[][Theorem\,7]{HKPV06} proved that sampling $\DPP(K)$ can be done in two steps:
        \begin{enumerate}
            \item draw $B_i\sim\Ber(\lambda_i)$ independently and denote $\lrcb{i_1,\dots,i_{N}} = \lrcb{i:B_i=1}$,\label{enum:sampling_step1}
            \item sample from the DPP with kernel $\tilde{K}(x,y) = \sum_{n=1}^{N}\phi_{i_n}(x) \phi_{i_n}(y)$.\label{enum:sampling_step2}
        \end{enumerate}
        In other words, all DPPs are mixtures of \emph{projection} DPPs, that are parametrized by an orthogonal projection kernel.
        In a nutshell, Step\,\ref{enum:sampling_step1} selects a component of the mixture and Step\,\ref{enum:sampling_step2} generates a sample of the projection $\DPP(\tilde{K})$.
        \mycitet[][Algorithm\,18]{HKPV06} provide a generic projection DPP sampler that we briefly describe.
        First, the projection DPP with kernel $\tilde{K}$ has exactly $N=\rank \tilde{K}$ points, $\mu$-almost surely.
        Then, the sequential aspect of the chain rule applied to sample $(X_1,\dots,X_N)$ with probability distribution
        \begin{equation}
        \label{eq:joint_distribution}
        \!
        \small
        \frac{\det\!\big[\tilde K(x_p,x_n)\big]_{p,n=1}^N}{N!}
            \prod_{n=1}^N \mu(\diff x_n)
        =
                \frac{\|\Phi(x_1)\|^2}{N}
            \mu(\diff x_1)
            \prod_{n=2}^{N}
                \frac{\dist^2\!\big(\Phi(x_n), \Span\lrcb{\Phi(x_p)}_{p=1}^{n-1}\!\big)}{N-(n-1)}
                \mu(\diff x_n),
        \end{equation}
        can be discarded to get a valid sample $\lrcb{X_{1}, \dots, X_{N}} \sim \DPP(\tilde{K})$.
        To each $x\in\bbX$ we associate a \emph{feature vector}
        $\Phi(x) \triangleq \lrp{\phi_{i_1}(x),\dots,\phi_{i_N}(x)}$,
        so that
        $\tilde{K}(x,y) = \Phi(x)^{\top} \Phi(y)$.

        A few remarks are in order.
        First, the LHS of \Eqref{eq:joint_distribution} defines an exchangeable probability distribution.
        Second, the successive ratios that appear in the RHS are the normalized conditional densities (w.r.t.\,$\mu$) that drive the chain rule.
        The associated normalizing constants are independent of the previous points.
        The numerators can be written as the ratio of two determinants and further expanded with Woodbury's formula.
        They can be identified as the incremental posterior variances in Gaussian process regression with kernel $\tilde{K}$ \citep[Equation 2.26]{RaWi06}.
        Third, the chain rule expressed in \Eqref{eq:joint_distribution} has a strong Gram-Schmidt flavor since it actually comes from a recursive application of the base$\times$height formula.
        In the end, DPPs favor configuration of points whose feature vectors $\Phi(x_1),\dots, \Phi(x_N)$ span a large volume, which is another way of understanding repulsiveness.
        The previous sampling scheme is exact and generic but, except for projection kernels, it requires the eigendecomposition of the underlying kernel.

        In the finite setting, this corresponds to an initial $\calO(M^3)$ cost, then the complexity of drawing exact samples is of order $\calO(M\Expe{|\calX|}^2)$
        \citep[see, e.g.,][]{Gil14,TrBaAm18}.
        Besides, there exist some alternative exact samplers.
        \citet{Pou19} and \mycitet{LaGaDe18} use a $\calO(M^3)$ Cholesky-based chain rule on sets; each item in turn is decided to be excluded or included in the sample.
        \mycitet{DeCaVa19} first sample an intermediate distribution and correct the bias by thinning the intermediate sample (with size smaller than\,$M$) using a carefully designed DPP.
        In certain regimes, this procedure may be more practical with an overall
        $\calO(M\operatorname{poly}(\Expe{|\calX|})\operatorname{polylog}(M))$ cost.
        In the continuous case, sampling exactly each conditional that appear in the right hand side of \Eqref{eq:joint_distribution} can by done by a rejection sampling mechanism with a tailored proposal.

        In applications where the costs related to exact sampling are a bottleneck, users rely on approximate sampling.
        Research has focused mainly on kernel approximation \citep{AKFT13} and MCMC samplers \citep{AnGhRe16, LiJeSr16c, GaBaVa17}.

        However, specific DPPs admit efficient exact samplers that do not rely on \Eqref{eq:joint_distribution}, e.g., uniform spanning trees \citep[UST,][Figure~\ref{fig:UST_kernel}]{PrWi98} or eigenvalues of random matrices.
        For instance, a $\beta$-ensemble is a set of $N$ points of $\bbR$ with joint distribution
        \begin{equation*}
        \setlength{\belowdisplayskip}{0pt}
        \setlength{\belowdisplayshortskip}{0pt}
        \setlength{\abovedisplayskip}{0pt}
        \setlength{\abovedisplayshortskip}{0pt}
        \label{eq:beta_ensemble_pdf}
        \frac{1}{Z_{N,\beta}}
        \prod_{p< n}
            \lrabs{x_p-x_n}^{\beta}
        \prod_{n= 1}^N
            \omega(x_n)
            \diff x_n, \quad \text{where } \beta>0.
        \end{equation*}
        For some choices of the weight function $\omega$, the $\beta$-ensemble can be sampled by computing the eigenvalues of simple tridiagonal \citep{DuEd02} or quindiagonal random matrices \citep{KiNe04}.
        In particular, ($\beta=2$)-ensembles correspond to projection DPPs \citep{Kon05}.
        They are therefore examples of continuous DPPs that can be sampled exactly in $\calO(N^2)$ time, without rejection.
        Some of these ensembles are of direct interest to MLers.
        The \emph{Laguerre} ensemble, for instance, has $\omega$ be a Gamma pdf, and corresponds to the eigenvalues of the empirical covariance matrix of \iid Gaussian vectors, see Figure\ref{fig:LUE}.
        Finally, we mention that \DPPy\ also features an exact sampler of the  multivariate extension of the \emph{Jacobi} ensemble which has been central in recent results on faster-than-Monte Carlo numerical integration \citep{BaHa16,GaBaVa19,MaCoAm19}.



\section{The \DPPy\ toolbox} 
\label{sec:the_dppy_toolbox}

    \lstset{language=mypython}

    \DPPy\ handles Python objects that fit the natural definition of the corresponding DPPs; see also the documentation\footnoteref{fn:docs}and the corresponding \href{https://github.com/guilgautier/DPPy/tree/master/notebooks}{Jupyter notebooks}, which showcase \DPPy\ objects.
    For example, \lstinline!FiniteDPP(kernel_type="correlation", **{"K": K})!\!
    instantiates a finite $\DPP(\bfK)$.
    Its two main methods,
    \lstinline{.sample_exact()} and
    \lstinline{.sample_mcmc()}
    implement the different exact samplers and current state-of-the-art MCMC samplers.
    To sample $k$-$\DPP$s, the additional
    \lstinline{.sample_exact_k_dpp()} and
    \lstinline{.sample_mcmc_k_dpp()} methods are available.

\begin{figure*}[!hb]
        \vspace{-0.5em}
        \centering
        \subfigure[2D Jacobi ensemble]{
            \includegraphics[height=10.5em]{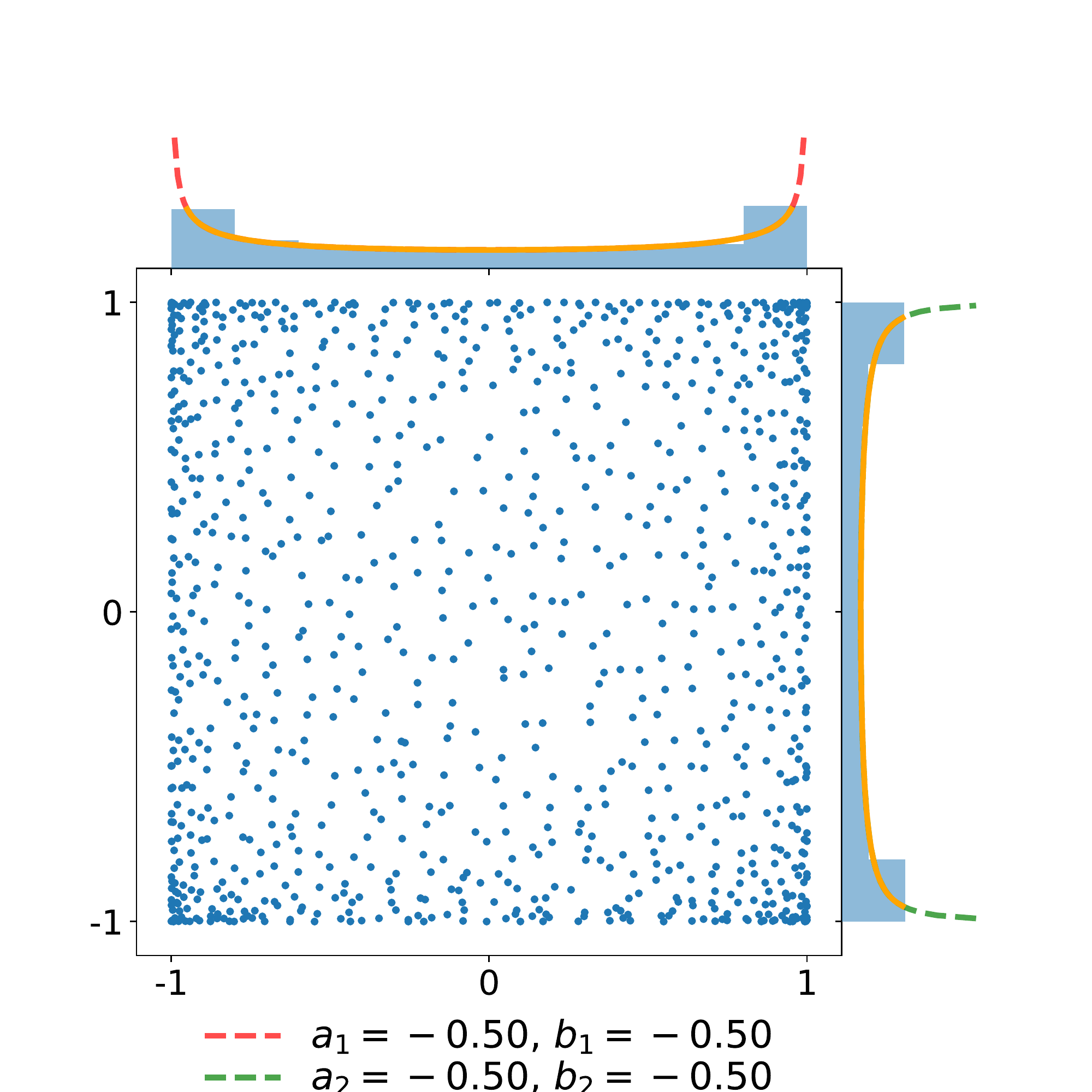}
            \label{fig:2D_jacobi_sample}
        }
        \subfigure[$\beta=2$-Laguerre ensemble]{
            \includegraphics[height=10.5em]{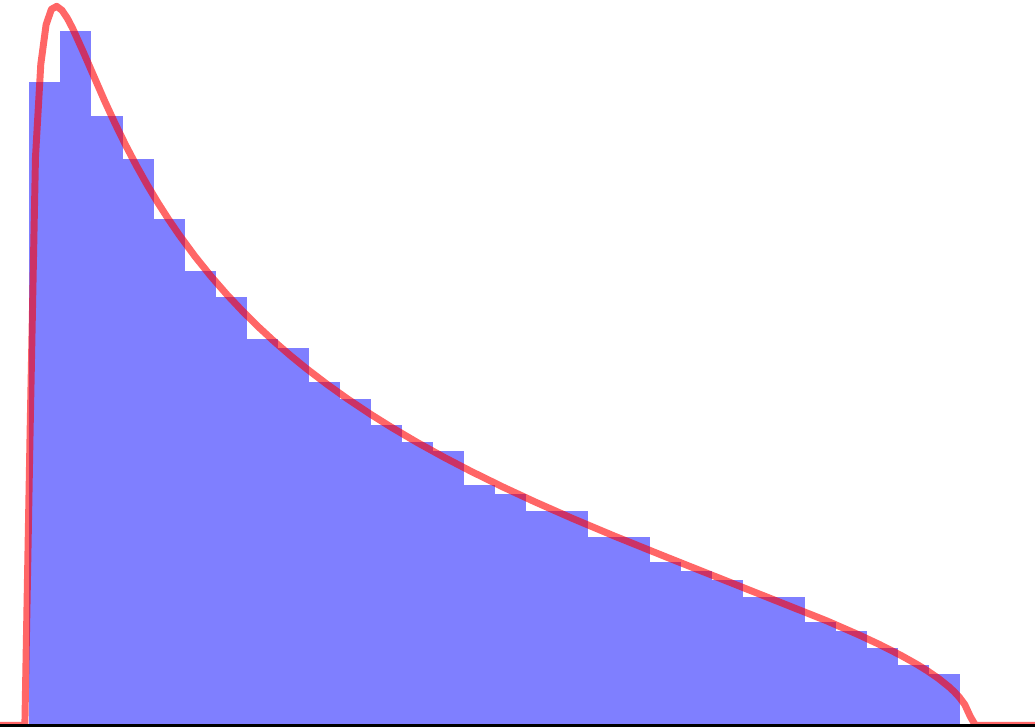}
            \label{fig:LUE}
        }
        \subfigure[$\bfK$ kernel of UST]{
            \includegraphics[height=10.5em]{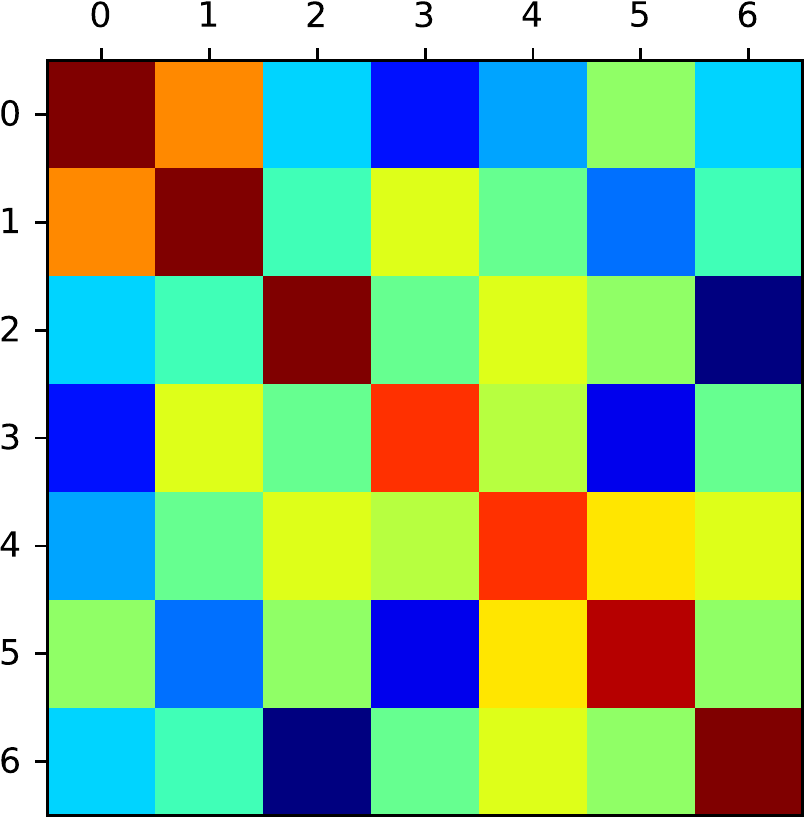}
            \label{fig:UST_kernel}
        }
        \vspace{-0.75em}
        \caption{Some displays available in \DPPy}
        \label{fig:DPPy_figs}
        \vspace{-10em}
    \end{figure*}

    A Laguerre $\beta$-ensemble is instantiated as
    \lstinline{LaguerreEnsemble(beta=2).}
    It can be sampled using either the full matrix model (eigenvalues of random covariance matrix) when $\beta\in\lrcb{1,2,4}$ with
    \lstinline{.sample_full_model()}
    or the tridiagonal one with
    \lstinline{.sample_banded_model()}, for $\beta > 0$.
    Samples can be displayed via
    \lstinline{.plot()} or
    \lstinline{.hist()} to construct the empirical distribution that converges to the Mar\v{c}enko-Pastur distribution, see Figure~\ref{fig:LUE}.

    \DPPy\ can readily serve as research and teaching support.
    \DPPy\ is also ready for other contributors to add content and enlarge its scope, \eg with procedures for learning kernels.

\clearpage

\acks{%
We acknowledge funding by European CHIST-ERA project DELTA, the French Ministry of Higher Education and Research, the Nord-Pas-de-Calais Regional Council, Inria and Otto-von-Guericke-Universit\"at Magdeburg associated-team north-European project Allocate, and French National Research Agency project BoB (n.ANR-16-CE23-0003).
}







\bibliography{biblio_new}

\end{document}